\documentclass{bmvc2k}
\usepackage{times}
\usepackage{amsmath}
\usepackage{amssymb}
\usepackage{cuted}
\usepackage{graphicx}
\usepackage{xcolor}
\usepackage{soul}
\usepackage{booktabs}
\usepackage{arydshln}
\usepackage{cleveref}
\usepackage{capt-of}
\usepackage{floatrow}
\usepackage{hyperref}
\usepackage{multicol}
\usepackage[toc,page]{appendix}
\usepackage{pifont}

\newcommand{\reb}[1]{{\textcolor{black}{#1}\xspace}}

\newcommand{\enc}{AVST\textsubscript{enc}}
\newcommand{\encmp}{AVST\textsubscript{enc-mp}}
\newcommand{\dec}{AVST\textsubscript{dec}}
\newfloatcommand{capbtabbox}{table}[][\FBwidth]
\title{\textsc{Audio-Visual Synchronisation in the wild}}

\addauthor{Honglie Chen}{hchen@robots.ox.ac.uk}{1}
\addauthor{Weidi Xie}{weidi@robots.ox.ac.uk}{1}
\addauthor{Triantafyllos Afouras}{afourast@robots.ox.ac.uk}{1}
\addauthor{Arsha Nagrani}{arsha@robots.ox.ac.uk}{1}
\addauthor{Andrea Vedaldi}{Vedaldi@robots.ox.ac.uk}{1}
\addauthor{Andrew Zisserman}{az@robots.ox.ac.uk}{1}
\addinstitution{
 Visual Geometry Group\\
 Department of Engineering Science\\
 University of Oxford\\
 Oxford, UK
}
\runninghead{CHEN ET AL.}{Audio-Visual Synchronisation in the wild}

\def\ie{\emph{i.e}\bmvaOneDot}

\def\eg{\emph{e.g}\bmvaOneDot}

\newcommand{\xmark}{\ding{55}}%

\usepackage[font={color=bmv@captioncolor,footnotesize}]{caption}

\begin{document}

\maketitle
\vspace{-3mm}
\begin{abstract}
In this paper, we consider the problem of audio-visual synchronisation applied to videos `in-the-wild' (\ie of general classes beyond speech).
As a new task, we identify and curate a test set with high audio-visual correlation, namely VGG-Sound Sync. 
We compare a number of transformer-based architectural variants specifically designed to model audio and visual signals of arbitrary length, while significantly reducing memory requirements during training.
We further conduct an in-depth analysis on the curated dataset and define an evaluation metric for open domain audio-visual synchronisation.
We apply our method on standard lip reading speech benchmarks, LRS2 and LRS3, with ablations on various aspects. Finally, we set the first benchmark for general audio-visual synchronisation with over 160 diverse classes in the new VGG-Sound Sync video dataset. 
In all cases, our proposed model outperforms the previous state-of-the-art by a significant margin. Project page: \url{https://www.robots.ox.ac.uk/~vgg/research/avs}

\end{abstract}
\vspace{-6mm}\section{Introduction}
\label{sec:intro}

In videos, the audio and visual streams are often strongly correlated,
presenting effective signals for self-supervised representation learning~\cite{arandjelovic18objects,Owens2018b}. 
A useful task in this area is audio-visual synchronisation, 
and several studies have shown promising results even without requiring any manual supervision~\cite{Chung16a,chung2019perfect,Afouras20b}. 
However, these works study this problem extensively on only one class --  human speech -- where even a slight offset is easily discernable. 

In this paper, 
rather than focusing on a specialised domain,
\eg human speech~\cite{Chung16,Chung16a,chung2019perfect,Afouras20b}, 
or videos with periodic sounds such as the tennis shots in a match~\cite{tennis}, 
we aim to explore audio-visual synchronisation on general videos in the wild (characterized by more than 160 sound classes). 
Solving this task would be extremely useful for a number of applications 
including video conferencing, television broadcasts and video editing,
which are largely done by `off-line' measurements or heavy manual processing~\cite{staelens2012assessing, 5332301,dassani2019automated}.


\begin{figure}[t]
\footnotesize
\centering
\includegraphics[width=.95\textwidth]{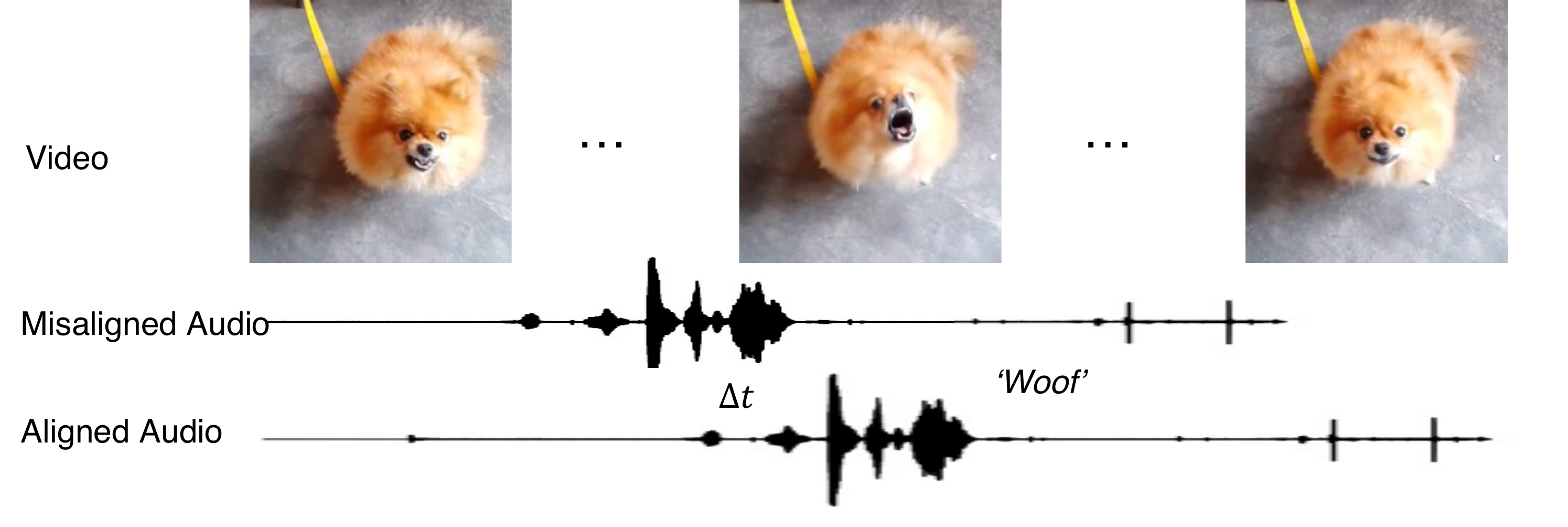}
\vspace{-.4cm}
\caption{\footnotesize \textbf{Audio-visual synchronisation in the wild}.
The goal of this work is to develop an audio-visual synchronisation method that performs well on general videos in-the-wild.
Unlike speech videos, highly correlated audio and visual events for general classes may occur briefly in the video (e.g.\ the bark of the dog in the centre of this clip). A short clip sampled randomly from the video might miss this fleeting moment; with longer input videos this becomes less probable.
With this in mind, we propose a Transformer based architecture that can operate on long sequences, and is able to perform audio-visual synchronisation on videos of ~160 general sound classes. \vspace{-.5cm}}
\label{fig:teaser}
\end{figure}

There are several challenges in automatic audio-visual  synchronisation for general classes. 
First, unlike the task of synchronising speech~\cite{nagrani17voxceleb:,Afouras19,Chung17,Chung17a},
which contains audio-visual evidence from the lips most of the time, 
videos from general classes 
\reb{may contain uniform sounds~(\eg~airplane engine sound, 
electric trimmer), ambient sounds~(\eg~wind, water, crowds, traffic), 
or small object sound sources~(\eg~players in an orchestra, birds), 
which make synchronisation extremely challenging or even impossible;
Second, 
for categories with strong audio-visual evidence, 
localising such signals can also be difficult,
for example:
temporally, `dog barking' may happen instantaneously, 
as shown in Figure~\ref{fig:teaser},
and spatially, unlike in speech synchronisation where visual cues are largely localised to lip motions, 
in general videos
the entire frame must be processed to accommodate different object classes;
Third, due to the aforementioned challenges,
it is unclear how to evaluate the synchronisation in general classes.}

In order to address these issues, 
first, we curate a new benchmark for general audio-visual synchronisation called VGG-Sound Sync using a subset of  VGG-Sound~\cite{Chen20a}. Specifically, this is built by
selecting classes and video clips that potentially have audio-visual correlation, 
and removing those classes and video clips that don't, \eg~uniform, ambient sound;
Second, compared with previous works, 
we use substantially longer input video sequences, 
so that the chance of having a synchronised audio and video event in the input increases. We explore several variants of Transformer-based architectures that can elegantly deal with these long sequences of variable lengths,
and that use self-attention to implicitly pick out the relevant parts in both space and time. 
Finally,
we conduct a thorough study on the VGG-Sound Sync test set, 
estimating the chance of audio-visual synchronisation for different clip lengths, and also define a set of metrics for evaluation.

Concretely, in this paper, 
we consider the problem of audio-visual synchronisation applied to `in-the-wild' videos, \ie~general classes beyond speech.
We make the following contributions: 
\reb{
(i) we identify and curate a subset of general classes from VGG-Sound, namely VGG-Sound Sync,
with potentially high audio-visual correlation;
(ii) we introduce a set of transformer-based architectures for audio-visual synchronisation, which can exploit the spatial-temporal correlations between audio and visual streams, 
such models can train and predict on variable length video sequences;
(iii) we conduct an analysis on the VGG-Sound Sync test set, 
and define an evaluation metric for audio-visual syncrhonisation on these videos;
(iv) we achieve state-of-the art synchronisation performance on standard lip reading speech benchmarks, LRS2, LRS3; and more importantly, set the first benchmark for audio-visual synchronisation in general (non-speech) classes. }

\vspace{-.6cm}
\section{Related Work}\label{sec:related}

\par{\noindent \bf Audio-visual synchronisation.}
Early works studied audio-visual synchronisation in talking faces~\cite{hershey1999audio,Slaney2000} using handcrafted features and statistical models.
~\cite{Chung16} developed a model for 
synchronizing lip movements to audio speech, based on a dual-encoder architecture trained with contrasting learning.
Follow-up works improved this pipeline by moving to noise-contrastive objectives~\cite{chung2019perfect}, or directly inferring the audio-visual offset conditional on cross-similarity patterns~\cite{Kim2021EndToEndLS}.
Lip synchronisation is an important component for pipelines
used for various visual speech related tasks, such as 
lipreading~\cite{Chung16b,Afouras19}, active speaker detection~\cite{Chung16} and sign language recognition~\cite{Albanie20}.
Although these works demonstrate strong synchronisation performance, they are limited in terms of deployment as they are applicable only on videos that include speech.
Our method generalizes to broader sound source classes and conditions, while also outperforming these works in the speech domain. 
Other closely related works have investigated lip-syncing~\cite{Halperin2019DynamicTA}, i.e.\ the temporal alignment of video and speech clips from different sources, speech-conditioned face animation~\cite{Chung17b,vougioukas2019realistic}, and audio-visual dubbing~\cite{Prajwal19lipsync,  yang2020large}.
Audio-visual synchronisation has been also used as a pre-text task for learning
general visual and audio representations~\cite{Owens2018b, Korbar18, patrick20multi-modal,Afouras20b,Cheng2020LookLA}. 
~\cite{khosravan2018attention} investigate the use of attention for audio-visual synchronisation on speech data.
~\cite{tennis} train models to detect synchronisation errors based on mismatch of event detection between the audio and visual stream.
~\cite{Casanovas2014AudiovisualEF} propose a method for  synchronising audio-visual recordings of the same events from different cameras.
Unlike the works above which use simple concatenation between audio and visual features, we employ encoder-based and decoder-based Transformers to implicitly match the relevant parts. 
\\[-6pt]

\par{\noindent \bf Audio-visual learning.} 
Our work is more broadly related to various works on audio-visual learning,
including audio-visual event detection~\cite{tian2018audio,Lin19dual},
sound-source localization~\cite{Arandjelovic16,qian2020multiple,Xu2020CrossModalRN,Afouras20b,gan2019self}, 
\reb{representation learning~\cite{nagrani2018learnable,alwassel2019self,asano2020labelling}}, audiovisual fusion~\cite{xiao2020audiovisual,jaegle2021perceiver,nagrani2021attention} and sound source separation~\cite{zhu2021vslowfast,zhu2021leveraging,zhao2018sound, Gao18graum,tzinis2021into}. \reb{More recently,~\cite{zhao2019sound} proposed to leverage temporal motion information to separate musical instrument sound.~\cite{Gan20music} further improved the sound separation models with explicit keypoint-based representations.  Another line of work explored audio synthesis using visual input: \cite{FoleyMusic2020} utilized body keypoints to synthesize music from a silent video, and~\cite{Koepke20} synthesized piano music from overhead views of the hands.~\cite{gao2019visual} converted monaural audio into binaural audio by injecting visual spatial information. }\\[-6pt]

\par{\noindent \bf Transformers.} 
Transformers~\cite{vaswani17attention} were originally introduced for NLP tasks,
in particular machine translation where they showed improvement over recurrent-based encoder-decoder architectures.
Since then they have been widely applied to a great range of problems,
including speech recognition~\cite{conformer}, language modelling~\cite{devlin18bert,dai2019Transformer}, 
object detection~\cite{carion2020endtoend,yao21efficientdetr}.
Recent works have even extended their use to visual feature extraction, replacing CNNs,
for \reb{classification~\cite{dosovitskiy2021an}}, 
semantic segmentation~\cite{Wu20visual,dosovitskiy2021an} and
video representation learning~\cite{bertasius2021spacetime}.
\reb{In the multi-modal domain,  
~\cite{Lin_2020_ACCV,tian2020avvp} explored unimodal and cross-modal temporal contexts simultaneously to detect audio-visual events, and~\cite{lee2021parameter} alleviated the high memory requirement of a vanilla Transformer by sharing the weights across layers and modalities. Audio-visual fusion using transformers has also been explored by new architectures such as Perceiver~\cite{jaegle2021perceiver} and MBT~\cite{nagrani2021attention}. 
}
\vspace{-2mm}
\section{Method}\label{sec:method}
\vspace{-2mm}
In this section, we describe our proposed method, which we call Audio-Visual Synchronisation with Transformers (AVST).
Our goal is to detect audio-visual synchronisation without the use of any manual annotation. 
Similar to prior work~\cite{Chung16,Owens2018b,Afouras20b}, we first use CNN encoders to extract visual and audio representations from unlabelled video (described in section~\ref{sec:rep}).
In \cref{sec:sync_modules}, we introduce three variants of our Transformer-based module that can jointly process visual and audio features, 
and discuss the pros and cons for each architecture. Finally in \cref{sec:loss}, we describe the contrastive learning objective used to train the model.
An overview of our architecture can be found in Figure~\ref{fig:model}.

\vspace{-4mm}
\subsection{Architecture}

\subsubsection{Audio and visual representations} \label{sec:rep}
The proposed model has two input streams, one ingesting a short video clip
$v_i\in \mathbb{R}^{3\times T \times H_v \times W_v}$ consisting of $T$ visual frames and
the other taking in an audio spectrogram $a_j\in\mathbb{R}^{1\times H_a \times W_a}$,  
where $i,j$ index the source of each modality~(e.g.~when $i=j$ the visual and audio signals come from the same video and are temporally aligned).
We compute representations for each modality using functions $f(\cdot; \theta_1)$ and $g(\cdot; \theta_2)$, which in this case are instantiated using CNN encoders:\\[-15pt]
\begin{align}
V_i &= f(v_i; \theta_1), \quad V_i\in\mathbb{R}^{c\times t_v \times  h\times w} \label{e:videorep}
\\
A_j &= g(a_j; \theta_2), \quad A_i\in\mathbb{R}^{c \times t_a}\label{e:audorep}
\end{align}
Both representations $V_i$ and $A_j$ have the same number of channels $c$, 
which allows us to jointly model the input video and audio with cross-modal attention.

\vspace{-3mm}
\subsubsection{{Synchronisation module}}\label{sec:sync_modules}
The visual and audio representations are formulated into a sequence of tokens, and passed through a Transformer~\cite{vaswani17attention} consisting of $N$ layers.
We introduce three variants of AVST, each one with a slightly different design choice for modelling audio-visual information. \\[-6pt]

\noindent\textbf{Encoder variant (\enc).}
The most straightforward step is to simply treat the dense visual features as a sequence of ``visual tokens''. 
To that end, the visual features are flattened over the spatial dimensions and concatenated to the audio features
after also prepending a learnable {\tt class} token~([\textsc{CLS}]), inspired by the BERT model~\cite{devlin18bert}.
In order for the model to distinguish the signals from the two modalities
and maintain spatio-temporal positional information (as all subsequent Transformer layers are permutation invariant), 
three types of encodings are also added to the audio and visual features:
modality encodings $E_{m} \in \mathbb{R}^{c \times 2}$, 
that indicate the type of feature (i.e.\ audio or visual);  
temporal encodings $E_{t_{\{v,a\}}} \in \mathbb{R}^{c \times t_{\{v,a\}}}$ 
and spatial encodings $E_{s} \in \mathbb{R}^{c \times h \times w}$, 
that keep track of absolute positions for the tokens:
{
\small
\begin{align}
&\overline{V_i} = \textsc{flatten}(V_i) + E_{m} + E_{t_v} + E_{s}, \\ \label{eqn:flatten}
&\overline{A_j} = A_j + E_{m} + E_{t_a},  \\
&Z_{ij} =  [[\textsc{CLS}];  \overline{V_i};  \overline{A_j}]
\end{align}
}%
where $[;]$ denotes a concatenation operation. 
The output result,
$Z_{ij} \in \mathbb{R}^{c \times (1+hwt_v+t_a)}$, 
is then fed into a Transformer Encoder~\cite{vaswani17attention},
that is composed of a stack of Multihead Self-Attention~(\textsc{MSA}), and feed forward networks (FFNs).
This module allows the tokens from both modalities to directly interact with each other through the self-attention operations: 
{
\small
\begin{align}
Y_{ij} = \textsc{Transformer-Encoder}( Z_{ij} ).
\end{align} 
}%

\begin{figure*}[t]
\footnotesize
\centering
\includegraphics[width=0.94\textwidth, height=0.40\textwidth]{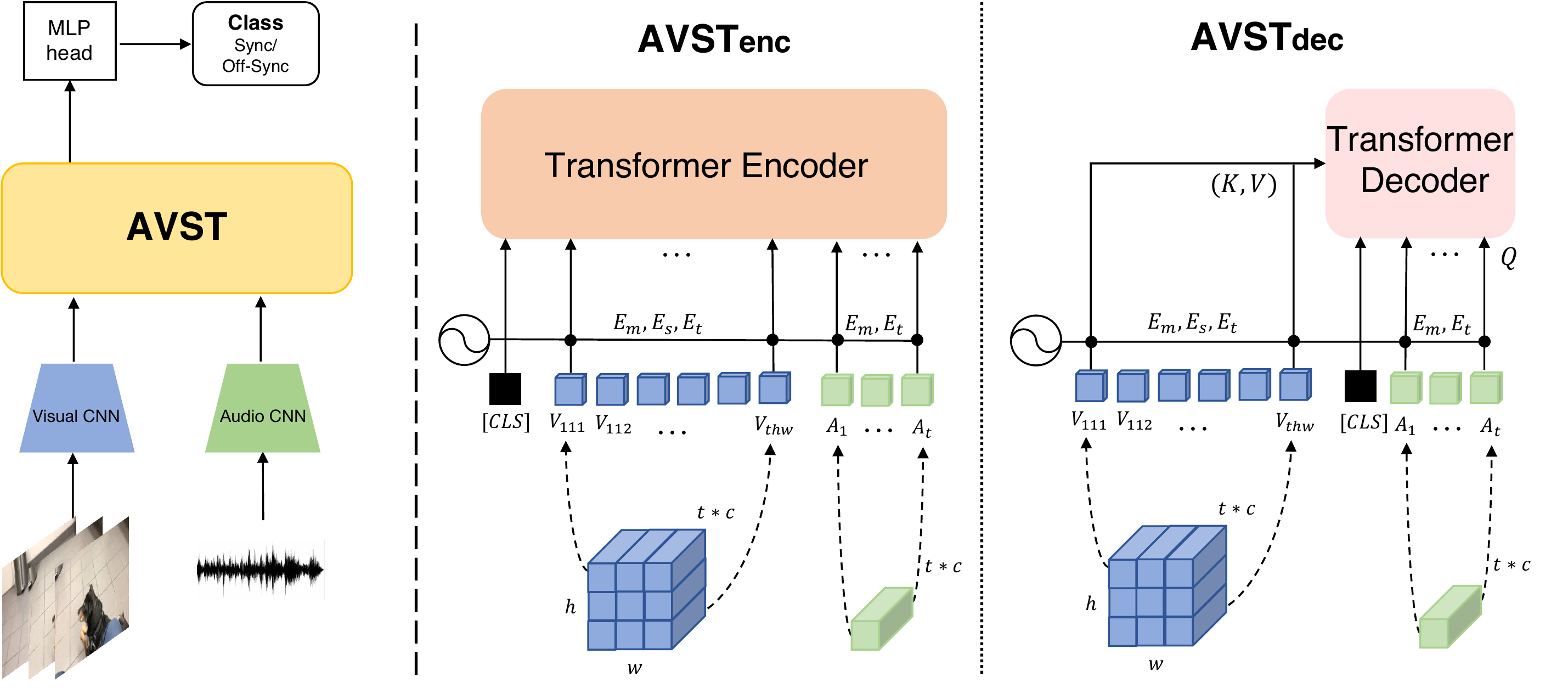}
\vspace{-3mm}
\caption{\footnotesize\textbf{The AVST model architecture and variants.} We use AVST to jointly model visual and audio representations computed from backbone CNNs, with an MLP head to predict audio-visual synchronisation (left). On the right, we show two variants of the AVST transformer backbone, the Encoder~(\enc) and Decoder variant~(\dec).
~\enc~uses self-attention for all audio and visual features, whereas in~\dec~the visual information is kept fixed, and the audio latents are used to \textsc{Query} the visual information which forms the \ \textsc{Key}, \textsc{Value} pairs.
\vspace{-3mm}}
\label{fig:model}
\end{figure*}

\vspace{3pt}
\par{\noindent \bf Max-pooled encoder variant (\encmp).} 
Naively feeding all visual features densely into the Transformer is computationally expensive, with a quadratic cost, $\mathcal{O}((hwt_v + t_a)^2)$,
which significantly limits the scalability to longer video sequences.
Thus, although the architecture is powerful, 
it heavily limits the audio-visual samples that can be processed in each batch, which in turn limits the number of negatives that can be used for training,
resulting in sub-optimal performance,
as we will show in the experiments~(section~\ref{sec:experiment}).

Rather than taking dense visual features as input, 
we propose a cheaper alternative, 
which consists of a simple Global Max Pooling (GMP) operation spatially on each frame. This reduces the length of the sequence that is input to the Transformer from $(hwt_v+t_a+1)$ to $(t_v+t_a+1)$; 
and thereby significantly lowers the memory footprint of the \textsc{MSA} module.
To obtain AVST\textsubscript{enc-mp} we simply replace the flattening operator in Equation~\ref{eqn:flatten} with GMP:
{
\small
\begin{align}
&\overline{V_i} = \textsc{GMP}(V_i) + E_{m} + E_{t_v} + E_{s}
\end{align}
}%

\noindent \textbf{Decoder variant  (\dec)}.
Using the visual feature from max-pooling is computationally efficient, 
however, it also removes spatial information in the visual representations, 
impairing the ability of audio features to probe fine-grained visual information, which may be required for certain general object categories.

To resolve the aforementioned challenge,
we consider an alternative architecture that uses
a Transformer decoder~\cite{vaswani17attention}, 
as shown in Figure~\ref{fig:model}~(right),
where dense visual features are kept fixed without self-attention and passed as the \textsc{Key} and \textsc{Value} inputs to every decoder layer, 
and audio features (concatenated along a [CLS] token, similarly to \enc) are passed as the \textsc{Query} inputs:
{
\small
\begin{align}
\textsc{Query} = \textsc{concat}([\textsc{CLS}], \overline{A}_j), \ \textsc{Key} = \textsc{Value} = \overline{V_i} \\
Y_{ij} = \textsc{Transformer-Decoder}(\textsc{Query}, \ \textsc{Key}, \ \textsc{Value})
\end{align} 
}%

\subsubsection{Output head}\label{sec:output_heads}
For all variants, we only use the first token~($Y_{ij}^1$),
of the output 
of the final encoder (or decoder) layer, 
corresponding to the [CLS] position in the input sequence.  
This functions as an aggregate representation of the whole output sequence and is fed to $h(\cdot; \theta_3)$, which we implement as an MLP head. The output is a synchronisation score that indicates to what degree the inputs $v_i$ and $a_i$ are in sync, $s_{ij} = h( Y_{ij}^1 ; \theta_3)$.

\subsection{Training objectives}\label{sec:loss}
Given mini-batches $\mathcal{B} = \{ (v_1, a_1), (v_2, a_2),..., (v_k, a_k)\}$ of temporally aligned audio-visual pairs,
the goal is to jointly optimize the entire pipeline in an end-to-end manner, 
so that the prediction scores for synchronised pairs $(v_i,a_i)$ are maximised, while the scores of out-of-sync pairs $(v_i,a_j)$ are minimised.
Training proceeds by minimising the commonly used  InfoNCE loss, defined as:\\[-15pt]
{
\small
\begin{align*}
\centering
\mathcal{L} = 
-\frac{1}{k}\sum_{i=1}^k
\left[ \log \frac{\exp(s_{ii})}{\sum_{j}\exp(s_{ij})} \right]
\end{align*}
}


\par{\noindent \bf Discussion.} 
Unlike previous works, which 
simply compute either the Euclidean distance or the cosine similarity between the audio and visual representations obtained from separate CNN streams to predict synchronisation, we use a Transformer model that jointly models the relationship between the audio and visual streams using attention over multiple layers. This is useful for attending to longer input sequences, where informative audio and video may only be localised in a short sub-sequence of the video.


\vspace{-3mm}
\section{Experiments}\label{sec:experiment}

In the following sections, we describe the datasets, evaluation
protocol and experimental details to thoroughly
assess our method.

\vspace{-3mm}
\subsection{Datasets}\label{sec:curatedata}
\par{\noindent \bf Audio-visual speech datasets:} 
We conduct experiments on two public audio-visual speech datasets,
namely, LRS2~\cite{Afouras19,Chung17,Chung17a} and LRS3~\cite{Afouras18d}, 
which have been created from British television footage 
and TED talks from YouTube respectively. 
Both datasets are distributed as short video clips of tightly cropped face tracks around the active speaker's head.
Since LRS3 is based on public YouTube videos, 
we also extract full-frame versions of the same clips for all splits (``pretrain'', ``trainval'' and ``test''). 
To distinguish between these two versions of LRS3, 
we refer to them as ``cropped'' and ``full-frame'' respectively. 
Note that for LRS2, only the ``cropped'' version is available.
\\[-8pt]

{\noindent \bf General sound dataset:}
Here, we construct a new benchmark called VGG-Sound Sync using a subset of VGG-Sound~\cite{Chen20a}, 
which was recently released with over $200k$ clips, and
each clip is labelled as one of the $300$ different sound categories. 
This dataset is conveniently audio-visual, 
in the sense that the object that emits sound is likely to be visible in the corresponding video clip. 
In the next section, we will detail the curation process.

\vspace{-2pt}
\subsection{Evaluation protocol} 
Depending on the downstream benchmarks, 
we consider two different evaluation protocols.
\vspace{-4mm}
\subsubsection{\reb{Audio-visual synchronisation on speech}}\label{subsec:speecheval}
For LRS2 and LRS3, 
we follow previous works and use an input of $5$ frames, 
extracted at 25fps.
During testing, the synchronisation scores were computed densely between each 5-frame video feature and all audio features in $\pm 15$ frame range.
Synchronisation was then determined to be correct if the lip-sync error was not detectable to a human, 
~\ie~the maximum score between two streams is  within $\pm 1$ frame~($\pm 0.04s$) 
from the ground truth~\cite{chung2019perfect}.
\vspace{-7mm}
\subsubsection{\reb{Audio-visual synchronisation on general classes}}
Compared to speech videos with audio-visual cues (the lip motion and  speech) spanning almost the entire clip,  
evaluating synchronisation on general videos potentially incurs two challenges:
(1) videos with only ambient or uniform sound, \eg~wind, wave, engine sound,
are unlikely to have any cues that can be used for synchronisation;
(2) the audio-visual cues for synchronisation are sometimes instantaneous, 
\eg~a dog barking may only last for less than 1s. 
In the following, 
we describe the evaluation benchmark and how it was constructed.\\[-8pt]

\vspace{-0.3cm}
\begin{table}[!htb]
\centering
\footnotesize
\begin{floatrow}
\begin{tabular}{lccccccc}
\toprule
Seconds 	& 1s 	 & 2s     & 3s     & 4s      & 5s     & 6s     \\\midrule
Audio-visual evident     	& $50\%$ & $56\%$ & $57\%$ & $62\%$  & $60\%$ & $59\%$ \\ 
Uniform/ambient sound 	& $30\%$ & $34\%$ & $35\%$ & $31\%$  & $35\%$ & $38\%$ \\ 
No sound/object & $20\%$ & $10\%$ & $8\%$ & $7\%$  & $5\%$ & $3\%$ \\ 
\bottomrule
\end{tabular}
\vspace{-1em}
\caption{\footnotesize Categorisation of video clips as duration of video varies.
\vspace{-0.3cm}}
\label{tab:noise}%
\end{floatrow}

\end{table}

{\noindent \bf Categorising video clips. \hspace{3pt}}
Here, we analyse the statistics of videos in the VGG-Sound test set,
by categorising each video clip into three classes, 
namely, audio-visual evident, uniform / ambient sound, 
missing sound / visual object.
Specifically,
we randomly sample 1200 video clips, where 
each clip is of different lengths between 1s and 6s
for manual verification.
As shown in Table~\ref{tab:noise}, 
the following phenomenon can be observed:
{\em First}, 
the proportion of clips with uniform or ambient sound remains roughly constant, 
as this error is caused by all the video clips of particular sound categories;
{\em Second}, 
as expected, with the increase of temporal lengths, 
the chance of having audio-visual cues for synchronisation increases.
Notably, when clips are over 2s, the error rate drops to around 10\%.

At this stage, we curate a subset of VGG-Sound by filtering the sound categories
to remove ones that are potentially dominated by uniform / ambient sound,
resulting in a test set of over 160 classes, 
95k training videos and 5k testing videos~(each of them lasts 10 seconds).\\[-8pt]

{\noindent \bf Verifying synchronisation of YouTube videos. \hspace{3pt}}
In this section, 
we conduct manual verification to serve two purposes:
{\em first}, 
as the video clips in VGG-Sound are all sourced from YouTube, 
their audio-visual alignments are not always guaranteed,
we aim to understand the chance of these videos being audio-visual synchronised, at least from the perspective of an ordinary human observer;
{\em second}, 
we aim to understand the human tolerance, by that we mean, 
how much temporal misalignment is noticeable for human observers.
In a practical evaluation, 
offsets smaller than such tolerance should be ignored or considered as correct.
In detail,
we randomly sample 500 example videos from our test set with 25fps,
and create 1000 audio-visual pairs, with each lasting 5s.
The temporal offsets between both streams vary from $[-0.8, +0.8]$ second,
for example, for one visual clip sampled at time $t$, 
its paired audio signal can be centered at any time between $t-0.8, t+0.8$,
we feed these pairs to human observers and ask a binary question:
{\em is the given audio-visual pair synchronised ?}
Please check the detailed statistics on proportions of videos considered to be syncd by a manual observer in Appendix~\ref{sec:appendixC}. \\[-8pt]


{\noindent \bf Summary. \hspace{3pt}}
To evaluate the synchronisation for videos of general classes,
we curate a test set from VGG-Sound, namely VGG-Sound Sync,
with ambient, uniform sound categories removed.
We only include audio-visual pairs of length between 2 - 6 second, 
that have a sufficiently high chance of containing informative cues for synchronisation. 
During evaluation, we decode the videos with 25fps, 
and construct audio or visual input by taking every 5th frame,
note that, this has the same effect as using input decoded from 5fps.
The synchronisation scores are computed for 
all audio-visual pairs with $[-15, -14, \dots, +14, +15]$ frame gaps.
Considering the challenging nature for audio-visual synchronisation in natural videos,
synchronisation is determined to be correct if the synchronisation error is not detectable by a human, i.e. the maximum score between two streams is within $\pm 5$ frames ($\pm 0.2s$) from the ground truth.


\vspace{-4mm}
\subsection{Implementation details}\label{sec.detail}
\noindent\textbf{Training curriculum.} 
Following prior work~\cite{Korbar18,Afouras20b}, 
we train our models in two stages:
in the first stage, 
we construct the mini-batches by sampling audio-visual clips from different videos, 
this provides easy (correspondence) negatives that helps the training converge. 
In the second stage, all the clips in a mini-batch are sampled from the same video, which provides harder (synchronisation) negatives.\\[-8pt]

\noindent\textbf{Training hyper-parameters.} 
On a P40 GPU with 24GB memory, 
we train \enc~with a batch-size of 4 (due to memory restrictions),
for \encmp~and \dec, we use a batch-size of 16 and 12 respectively, 
thereby allowing more negatives per batch. \\[-8pt]

\noindent\textbf{Architectural Details.} Unless otherwise specified, our Transformer encoder consists of 3 layers, 4 attention heads and a hidden unit dimension of 512.
Typically $H=W=224$ and $h=w=14$. 
We refer the readers to Appendix~\ref{sec:appendixA} for more details.

\vspace{-3mm}
\subsection{Results on speech datasets}
We first report experimental results on LRS2 and LRS3, and perform a number of ablations on different architectural design choices.
We also analyse the model's robustness on cases, 
where the visual or audio signal is partially unavailable. \\[-8pt]


\par{\noindent \bf Architectures comparison.}
To compare our proposed architecture variants and assess their trade-offs, we train and evaluate them on the ``full-frame'' version of LRS3 and show results of all three Transformer variants in Table~\ref{tab:results_lrs}. 
Due to the memory restrictions, 
we can only train \enc~ with a fixed length of 5 frames, 
whereas for the other two architectures, 
training is done with variable sequence length and larger batch size~(see section~\ref{sec.detail}).
We observe a large gap~(6\% -- 7\%) between the performance of ~\enc~and the other two variants, 
which indicates that \enc~suffers from the reduced number of negatives. 
We also note that~\dec~can localise sound sources 
because it preserves spatial information, 
but shows slightly worse performance than~\encmp~on speech datasets. 
We conclude that \encmp~ is a light-weight solution that offers the best performance when the sounding objects (\eg lips) are clear and unique, which need little fine-grained spatial information. \\[-8pt]

\par{\noindent \bf Comparison to the state-of-the-art.}
We compare our method to previous work on ``full-frame'' LRS3 in the top half of Table~\ref{tab:results_lrs}. 
We show a significant improvement compared to the AVobjects baseline~(16\% gain) on short input~(5 frames) reaching up to an almost saturated $98.6\%$ accuracy with 15 frames. 
In the bottom half of Table~\ref{tab:results_lrs}, we further summarise our results for experiments on the ``cropped'' LRS2 dataset. Here too, we observe that our method greatly outperforms both the SyncNet\cite{Chung16a} and PM~\cite{chung2019perfect} baselines, and achieves almost perfect accuracy with $15$ frames of input during test time. 

Since \encmp~ shows superior performance on speech datasets using a light-weight architecture, we conduct the rest of the analysis on speech data using~\encmp. In addition, in order to compare with SyncNet and PM, we use the same fixed length of 5 frames during training and testing.
\\[-8pt]
\par{\noindent \bf Number of Transformer Layers.}
We ablate the Transformer depth on the LRS2 dataset in Table~\ref{tab:depth}.
As more layers are added, the performance consistently improves, 
achieving the best performance with 3 layers. 
This confirms the effectiveness of self-attention in jointly modelling audio and visual information.\\[-8pt]

\par{\noindent \bf Robustness test.}
To mimic real-world scenarios,
where sound sources and their corresponding sound might not appear together at every frame,
we further conduct experiments to assess the robustness of our model on the LRS2 dataset by randomly masking input audio or video frames.
We mask 1 frame
for each or both modalities. As can be seen in Table~\ref{tab:robost}, 
we find that for short inputs this causes a significant performance drop, 
however with longer inputs, we achieve comparable results to the non-masked case in Table~\ref{tab:results_lrs}.

\begin{table*}[t]
\footnotesize
\setlength{\tabcolsep}{5pt}
\centering
\begin{tabular}{l c c c  c c c c c c}
\toprule

     & &  &  &  \multicolumn{6}{c}{Clip Length in frames (seconds)}    
\\ 
\cmidrule(lr){5-10}
Model & \# Params. & Var. & Dataset & 5(0.2s) &7(0.28s) & 9(0.36s) & 11(0.44s) & 13(0.52s) & 15(0.6s) \\ \midrule
AVobjects~\cite{Afouras20b} & 69.4M & \xmark & LRS3 & $61.8$ & $72.0$ & $79.7$ & $85.4$  & $89.5$ & $91.8$ \\ 
\enc &  42.6M &  \xmark & LRS3 & $70.2$ & $77.1$ & $83.3$ & $88.4$  & $92.0$ & $94.4$ \\
\dec &  44.5M & $\checkmark$ & LRS3  & $75.7$ & $86.4$ & $89.4$ & $94.0$  & $95.1$ & $96.9$ \\
{\bf \encmp}  &  42.4M & $\checkmark$ & LRS3  & {\bf 77.3} & {\bf 88.0} & {\bf 93.3} & {\bf 96.4}  & {\bf 97.8} & {\bf 98.6} \\ \midrule

SyncNet~\cite{Chung16a} &13.6M & \xmark & LRS2 & $75.8$ & $82.3$ & $87.6$ & $91.8$  & $94.5$ & $96.1$ \\ 
PM~\cite{chung2019perfect} &13.6M& \xmark & LRS2 &$88.1$ & $93.8$ & $96.4$ & $97.9$  & $98.7$ & $99.1$ \\  
{\bf \encmp}  &42.4M& $\checkmark$ & LRS2 & {\bf 91.9} & {\bf 97.0} & {\bf 98.8} & {\bf 99.6}  & {\bf 99.8} & {\bf 99.9} \\ \bottomrule 
\end{tabular}
\vspace{-.2cm}
\caption{{\bf Architecture comparison on LRS3 and LRS2.}
We use the `full-frame' dataset. 
`Var': whether models are trained and tested using variable length inputs. `5-15' refers to the number of input frames to corresponding models.}
\label{tab:results_lrs}
\end{table*}

\begin{table}[t]
\vspace{-3mm}

\footnotesize
\RawFloats
\setlength\tabcolsep{3pt}
\begin{minipage}{.49\linewidth}
\centering
\begin{tabular}{lccccccc}
\toprule
  &  \multicolumn{6}{c}{Clip Length~(frames)}    
\\ 
\cmidrule(lr){2-7}
\# Layers  & 5 	    & 7      & 9      & 11      & 13     & 15  

\\\midrule
1  & $89.1$ & $94.0$ & $96.8$ & $98.4$  & $99.1$ & $99.4$ \\ 
2 & $91.6$ & $95.4$ & $97.6$ & $98.8$  & $  99.1$ & $99.6$ \\ 
3  & $92.0$ & $95.5$ & $97.7$ & $98.8$  & $99.3$ & $99.6$ \\ 
\bottomrule
\end{tabular}
\vspace{8pt}
\caption{\footnotesize \textbf{Ablation on Transformer depth (LRS2).} Performance increases with depth.}
\label{tab:depth}
\end{minipage}\hfill
\begin{minipage}{.49\linewidth}
\centering
\begin{tabular}{lccccccc}
\toprule
  &  \multicolumn{6}{c}{Clip Length~(frames)}    
\\ 
\cmidrule(lr){2-7}
Mask	    & 5 	   & 7      & 9      & 11      & 13     & 15     \\\midrule
Audio   & $73.1$ & $85.3$ & $92.6$ & $96.1$  & $98.0$ & $99.2$ \\ 
Visual	& $76.5$ & $87.3$ & $93.4$ & $96.9$  & $98.2$ & $99.3$ \\ 
Both & $71.7$ & $84.0$ & $91.2$ & $95.6$  & $97.7$ & $99.1$ \\ 
\bottomrule
\end{tabular}
\vspace{8pt}
\caption{\footnotesize \textbf{Robustness test on LRS2.} 1 frame is masked during train and test.}
\label{tab:robost}
\end{minipage}
\vspace{-3mm}
\end{table}

\begin{table}[t]
\footnotesize
\RawFloats
\setlength\tabcolsep{2pt}
\begin{minipage}{.49\linewidth}
\centering
\renewcommand{\arraystretch}{1.5}
\begin{tabular}{lcccccccc}
\toprule
  &  \multicolumn{6}{c}{Clip Length in frames (seconds)}    
\\ 
\cmidrule(lr){2-6}
Method 	  					        & 10(2s)     & 15(3s)     & 20(4s)      & 25(5s)     & 30(6s)     \\\midrule
AVobjects~\cite{Afouras20b} & $37.2$ & $42.6$ & $45.1$  & $47.3$ & $49.4$ \\ 
\encmp                      & $39.0$ & $44.1$ & $46.8$  & $49.7$ & $51.8$ \\
\dec                        & $40.6$ & $44.6$ & $47.6$  & $50.4$ & $52.7$ \\

\bottomrule
\end{tabular}
\vspace{6pt}
\caption{\footnotesize  \textbf{Audio-visual synchronisation results on VGG-Sound Sync}. We show results for sequences of $10-30$ frames. Our model outperforms the state of the art AVObjects~\cite{Afouras20b} on sequence lengths $ \geqslant 10$. }
\label{tab:results_vgg}
\end{minipage}\hfill
\begin{minipage}{.499\linewidth}

\centering
\includegraphics[width=1\textwidth]{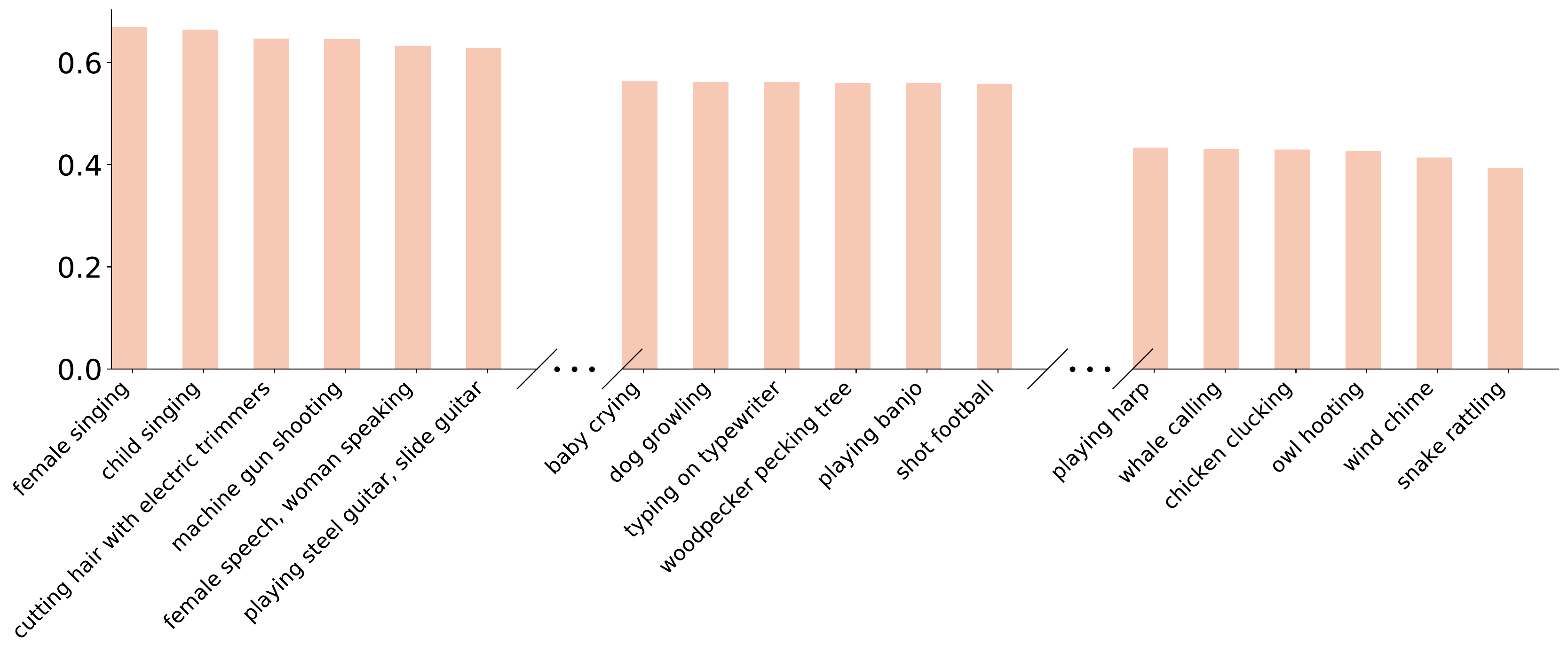}
\vspace{0.2em}
\captionof{figure}{\footnotesize  \textbf{Per-class accuracy on VGG-Sound Sync}.}

\label{fig:class_stat}
\end{minipage}
\end{table}

\vspace{-1mm}
\begin{figure*}[t]
\centering
\subfigure[\footnotesize Localise Visual Sound~\cite{Chen21}]{\includegraphics[width=.93\textwidth]{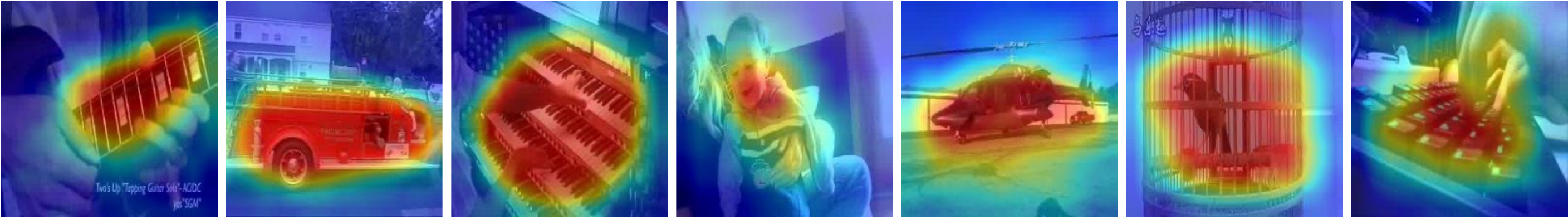}\label{Fig.lvs}}
\\[-2pt]
\subfigure[\footnotesize\vspace{-1.6cm}\dec]{\includegraphics[width=.95\textwidth]{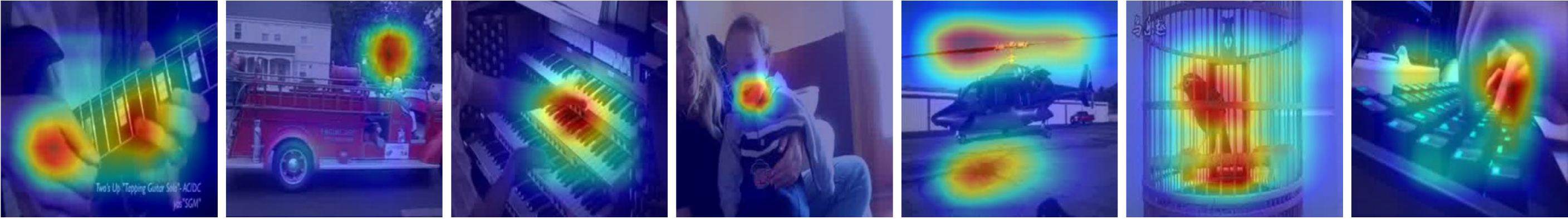}\label{Fig.dec}}

\vspace{-.4cm}
\caption{\footnotesize\textbf{Attention heatmaps on VGG-Sound Sync.} 
We compare the heatmaps that we obtain with the \dec~model to
the state-of-the-art method for sound source localization~\cite{Chen21}.  
It is interesting to note that while~\cite{Chen21} highlights discriminative parts of the objects that are generally associated with the sound and are therefore \textit{sufficient to identify it}  -- \ie the entire musical instrument, firetruck and helicopter -- our method focuses on the parts that exhibit some motion -- \ie the player's hands, the firetruck siren and the helicopter's rotor -- that \textit{modify or create sound} and are necessary to solve the much more challenging synchronisation task. 
\vspace{-.5cm}}
\label{fig:vgg_visual2}
\end{figure*}

\vspace{-4mm}
\subsection{Results on general sound classes}
In this section, we report audio-visual synchronisation results on the VGG-Sound~Sync dataset consisting of videos with general sound classes, 
and compare with several strong baselines. 
Results are provided in Table~\ref{tab:results_vgg}. 
\reb{First, while comparing with the recent AVobjects~\cite{Afouras20} method,
both of our models show superior results on all input lengths, 
this is because (1) we trained on variable input lengths, 
where longer samples contain richer audio-visual evidence; 
and (2) the use of Transformer based architectures~(\enc~and \dec) can implicitly discover the important temporal parts in long sequences.
Second, in contrast to the results in speech datasets~(Table~\ref{tab:results_lrs}), we note that \dec~has higher accuracy than \enc~on general videos. 
The reason is that general videos contain complex visual scenes and, 
compared to other variants, 
\dec~can extract fine-grained spatial information in such situation by explicitly computing the attention between image regions and the audio sequence, therefore showing better performance.}
Finally, 
we analyse the performance for each class of VGG-Sound Sync dataset in Figure~\ref{fig:class_stat}, and find that the performance is highly class dependant, with the best class~(`child singing') achieving 75.7\%, and most highly performing classes containing strong audio-visual correlations,~\eg `female singing',`playing steel guitar', etc. 

\vspace{-4mm}
\subsubsection{Visualisation of attention heatmaps}
We visualise the attention heatmaps of our model on samples from VGG-Sound Sync in Figure~\ref{fig:vgg_visual2}~(refer to Appendix~\ref{sec:appendixD} for more qualitative results).
For general object classes in VGG-Sound Sync, 
the model manages to pick up on interesting sources of motion that produce sound. 
When comparing the heatmaps produced by \dec~to the current 
{\em state-of-the-art} sound source localization method~\cite{Chen21},
we notice that our method attends to regions that \textit{create or modify} sound, \eg hands, lips, helicopter rotor, etc, while~\cite{Chen21} tends to localise the entire object.


\vspace{-4mm}
\section{Conclusion}\label{sec:conclusion}
We revisit the problem of audio-visual synchronisation in human speech 
and introduce a new general class audio-visual synchronisation benchmark called VGG-Sound Sync. 
We experiment with different variants of Transformer-based architectures,
analyse several critical components, 
and conduct thorough ablation studies to validate their necessity.
Consequently, our proposed architecture 
sets new {\em state-of-the-art} results on LRS2 and LRS3, 
and provides baselines for general sound audio-visual synchronisation.


\subsection*{Acknowledgements}
This work is supported by the UK EPSRC CDT in Autonomous Intelligent Machines and Systems, the
Oxford-Google DeepMind Graduate Scholarship, the Google PhD Fellowship, the EPSRC Programme Grant VisualAI EP/T028572/1, and a Royal Society Research Professorship.

{\small \bibliography{shortstrings,vgg_local,vgg_other,refs,vedaldi_specific,vedaldi_general}}
\clearpage
\begin{appendices}
\section{Implementation details}\label{sec:appendixA}
Here, we describe the architecture details, 
and hyper-parameters used during training.\\[-6pt]

\noindent\textbf{Input Features.}
We follow previous works and use an input of $5\sim15$ frames in LRS2 and LRS3 extracted at 25 FPS, 
and use $5\sim30$ frames in VGG-Sound Sync dataset with 5 FPS. 
Visual frames are resized to a $224  \times 224  \times 3 \times T$ tensor without cropping, indicating height, width, channel, frames.
The visual feature map before max-pool/reshape has a dimension of $14 \times 14$.
For the visual stream, we use a ResNet18 2D+3D as backbone. For instance,  an input of $224  \times 224  \times 3 \times 5$ would result the output visual feature a dimension of $14  \times 14  \times 512 \times 5$.
Audio spectrograms are extracted using an FFT with window size of 320 and hop length 40.
A lightweight VGG-M is then used to process the audio spectrogram giving tensors of dimension $512 \times T$. Encoders are initialised from scratch.\\[-6pt]

\noindent\textbf{Training hyperparameters.} 
We use a learning rate of $1e^{-4}$, Adam optimiser, trained for 100 epochs on 2 P40 GPUs.\\
\vspace{-15pt}
\section{Robustness test}
The aim of this experiment is to mimic real-world scenarios, as sound and its corresponding sound source might not happen at the same time. We mask random $n$ frame length of input for one or both modalities, \ie replace video frames or audio segments with zeros, we show results here when more than 1 frame is masked. With short input sequence,~\eg 5 frames, the performance drops significantly as the mask length increases,~\eg 3 frames, however, as the input length increases, our model shows robust performance at 15 frames. 
\begin{table}[!htb]
\footnotesize
\centering
\begin{tabular}{lccccccc}
\toprule
&    &  \multicolumn{6}{c}{Clip Length in frames (seconds)}    
\\ 
\cmidrule(lr){3-8}

Mask modality	& $\#$ Mask Length & 5(0.2s) &7(0.28s) & 9(0.36s) & 11(0.44s) & 13(0.52s) & 15(0.6s)     \\\midrule
Audio   &   1 Frame  &$73.1$ & $85.3$ & $92.6$ & $96.1$  & $98.0$ & $99.2$ \\ 
Visual	&   1 Frame  &$76.5$ & $87.3$ & $93.4$ & $96.9$  & $98.2$ & $99.3$ \\ 
Both 	&   1 Frame  &$71.7$ & $84.0$ & $91.2$ & $95.6$  & $97.7$ & $99.1$ \\ 
Both 	&   2 Frames  &$66.8$ & $83.2$ & $91.0$ & $95.5$  & $97.5$ & $99.1$ \\ 
Both 	&   3 Frames  &$56.1$ & $79.6$ & $90.3$ & $95.2$  & $97.2$ & $99.0$ \\
\bottomrule
\end{tabular}
\vspace{8pt}
\caption{\footnotesize \textbf{Robustness test on LRS2.} As the input sequence length increases, even when we mask more frames, the performance remains robust.}
\label{tab:depth}
\end{table}

\section{Synchronisation on general sound classes}\label{sec:appendixC}

\par
\begin{figure}[t]
\centering
\includegraphics[width=0.6\textwidth]{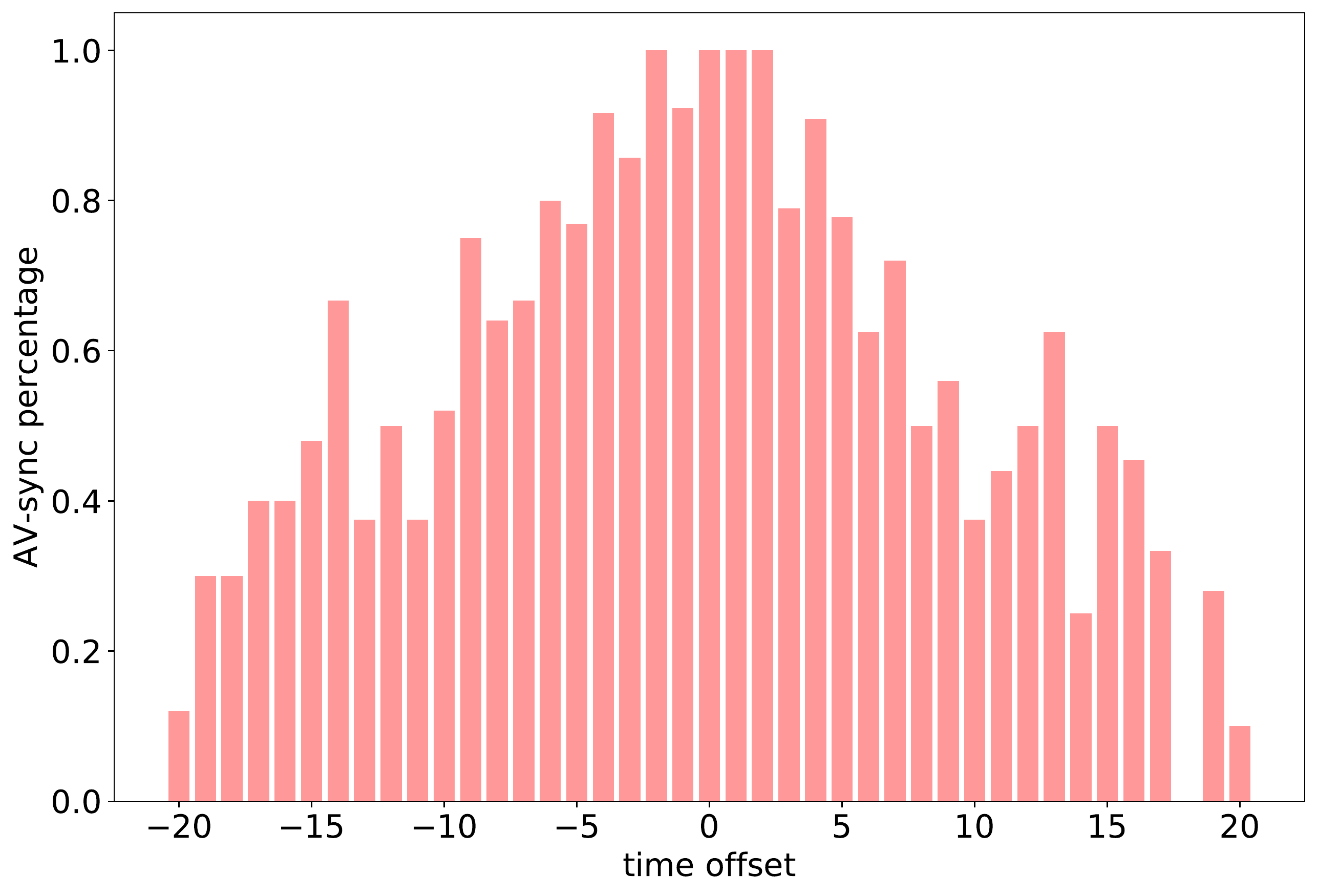}
\caption{\footnotesize Proportions of videos considered to be syncd by a manual observer,  Histogram demonstrating if the object-sync error is detectable for different offset values.}
\label{fig:histo}
\end{figure}

{\noindent \bf Human tolerance on general classes.}
Here, we analyse the synchronisation tolerance on general classes by manually creating misaligned audio-visual clips and asking human observers to verify whether the given clip is synchronised. Specifically, we randomly sample 500 videos from all classes (160 classes) in VGG-Sound Sync. The misaligned audio-visual pairs are then generated by randomly choosing a offset within $\pm 20$ frames~(0.8s). During manual verification, we create 25 audio-visual pairs for each offset, with 1000 audio-visual pairs in total.
As shown in Figure~\ref{fig:histo}, 
all sample videos are manually verified as synchronised at time offset 0, indicating the downloaded videos contain high-quality audio-visual synchronisation naturally.
In addition, an error below $\pm 5$ frames~(0.2s) is indistinguishable for human (Approximately, $90\%$ of the videos are verified as synchronised within $\pm 5$ frames). Furthermore, the proportions of videos considered to be synchronised decrease significantly beyond such offsets.
Therefore, we allow a prediction offset of up to $\pm 5$ frames during evaluation for general sound classes. 

\vspace{5pt}
\par
\begin{figure}[!htb]
\centering
\includegraphics[width=.98\textwidth]{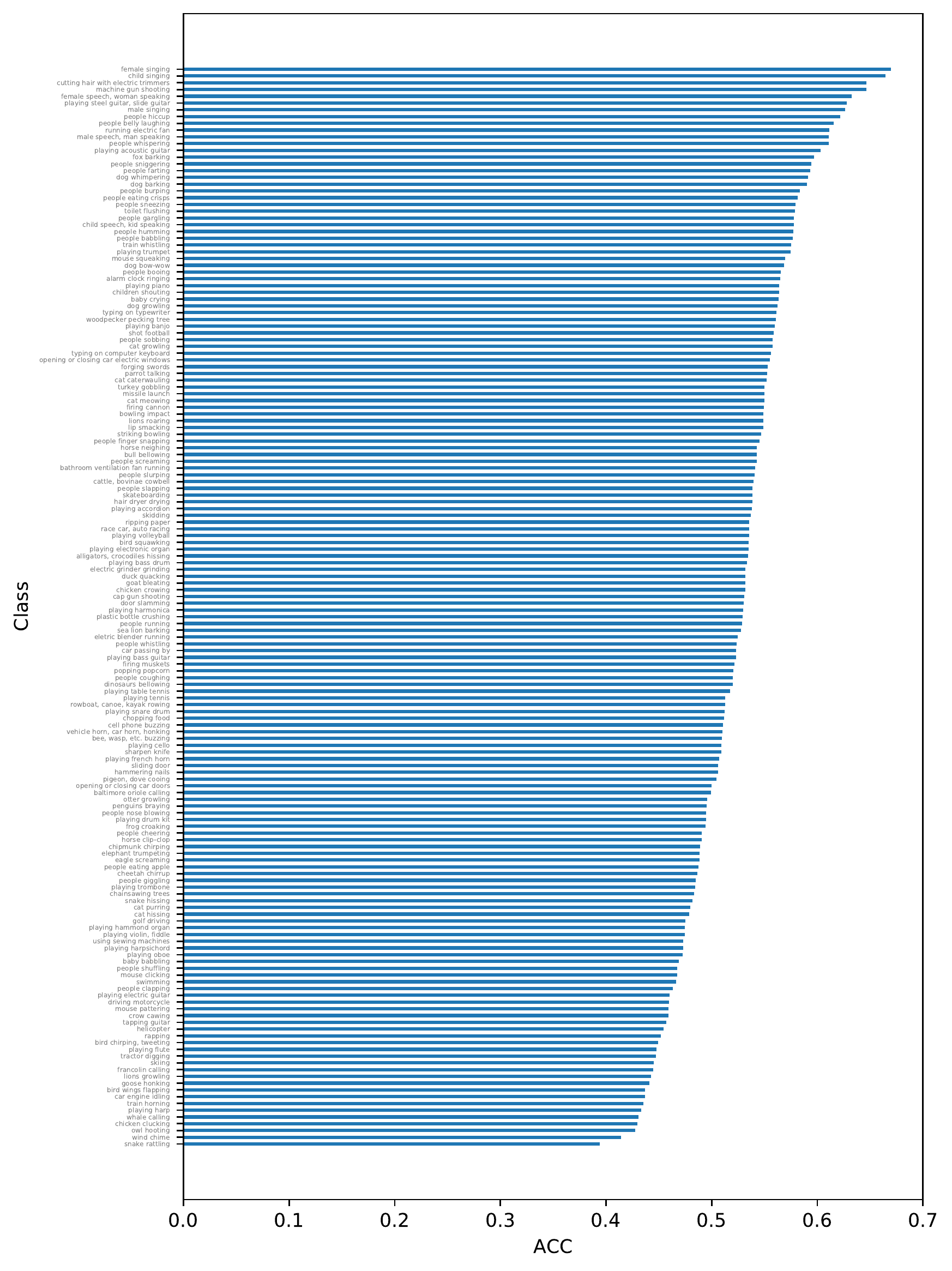}
\caption{Per class accuracy in VGG-Sound Sync.}
\vspace{-5mm}
\label{fig:class_stat_full}
\end{figure}

{\noindent \bf Per-class synchronisation accuracy.}
Please refer to Figure~\ref{fig:class_stat_full} for per class accuracy in VGG-Sound Sync dataset. We list the 160 classes we select from VGG-Sound in the end of this document.

\vspace{5pt}
\par
\begin{figure*}[!htb]
\subfigure[\textbf{Bull bellowing.} In this case, the bull makes sound only in the first few seconds, although visual information exist over the entire clip, the missing audio causes the low score.]{\label{fig:a}
\includegraphics[width=.96\textwidth]{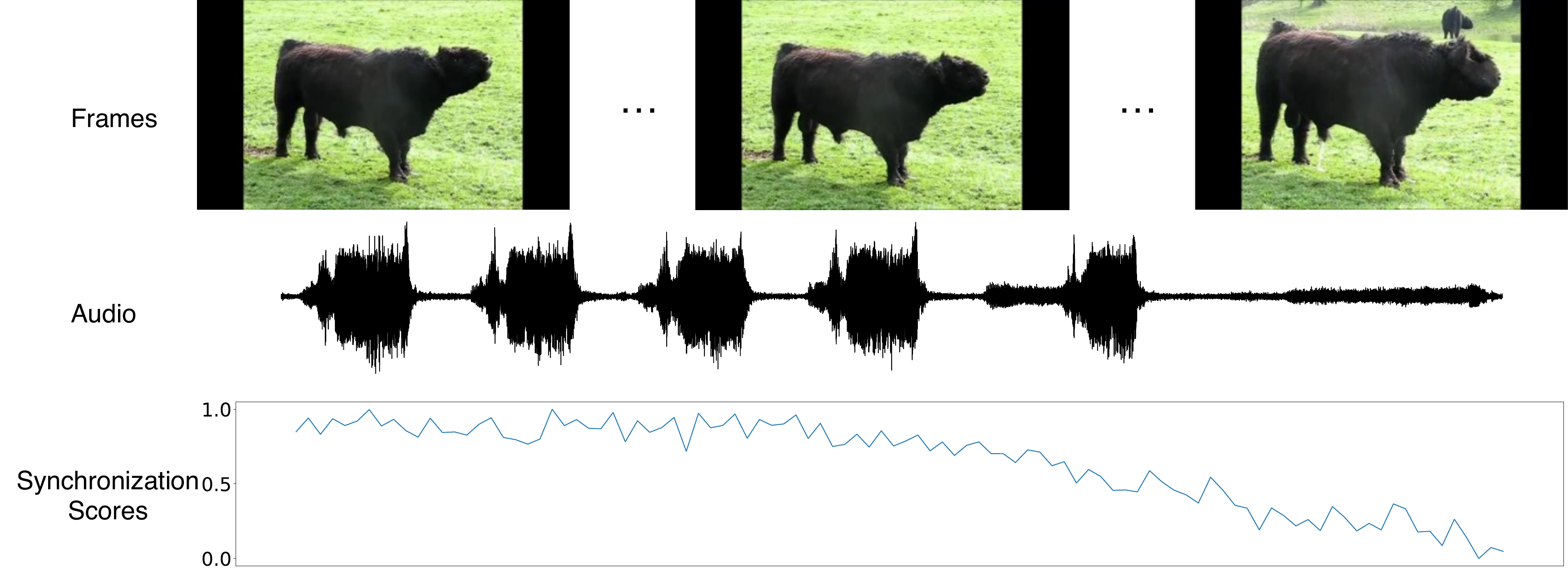}}\vspace{5mm}
\subfigure[\textbf{Playing violin, fiddle.} On the other hand, this example contains violin sound throughout the video clip, however, the violin player is not present all the time, causing the low score.]{\label{fig:b}
\includegraphics[width=0.96\textwidth]{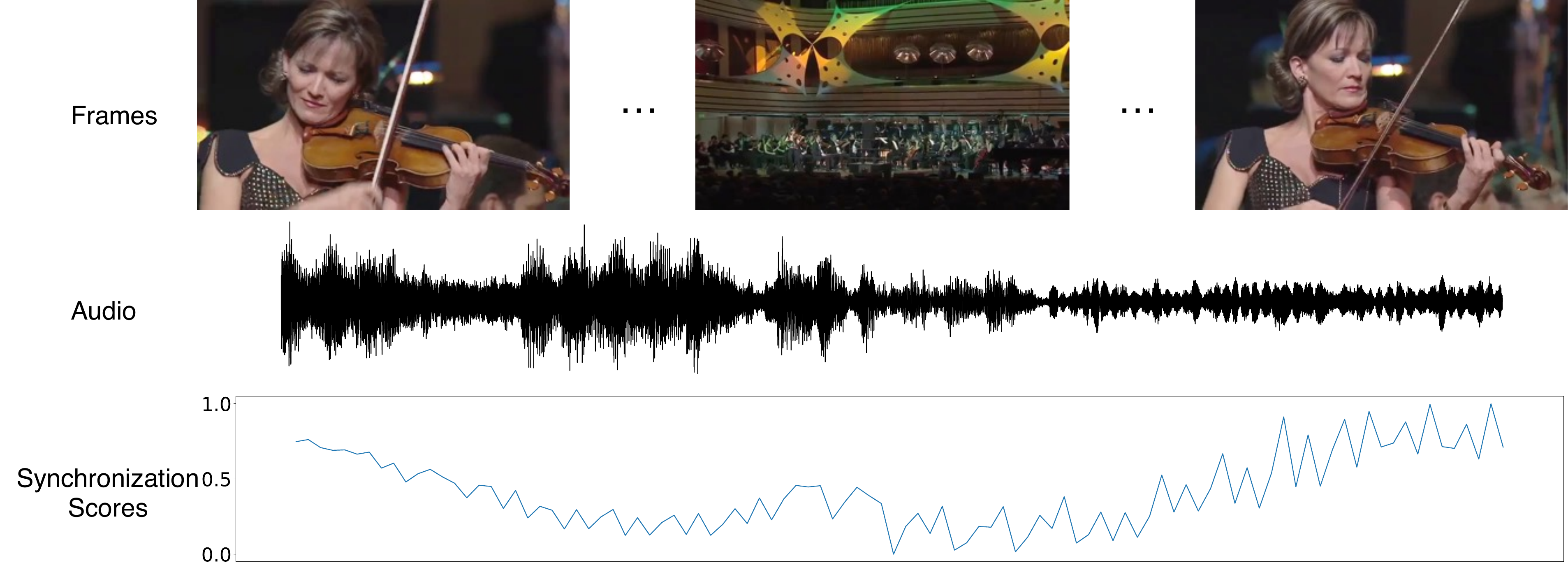}}\vspace{5mm}
\subfigure[\textbf{Electrical trimmer.} In this example, the last few seconds contain no information for both modalities, therefore the synchronisation score is extremely low.]{\label{fig:c}
\includegraphics[width=0.96\textwidth]{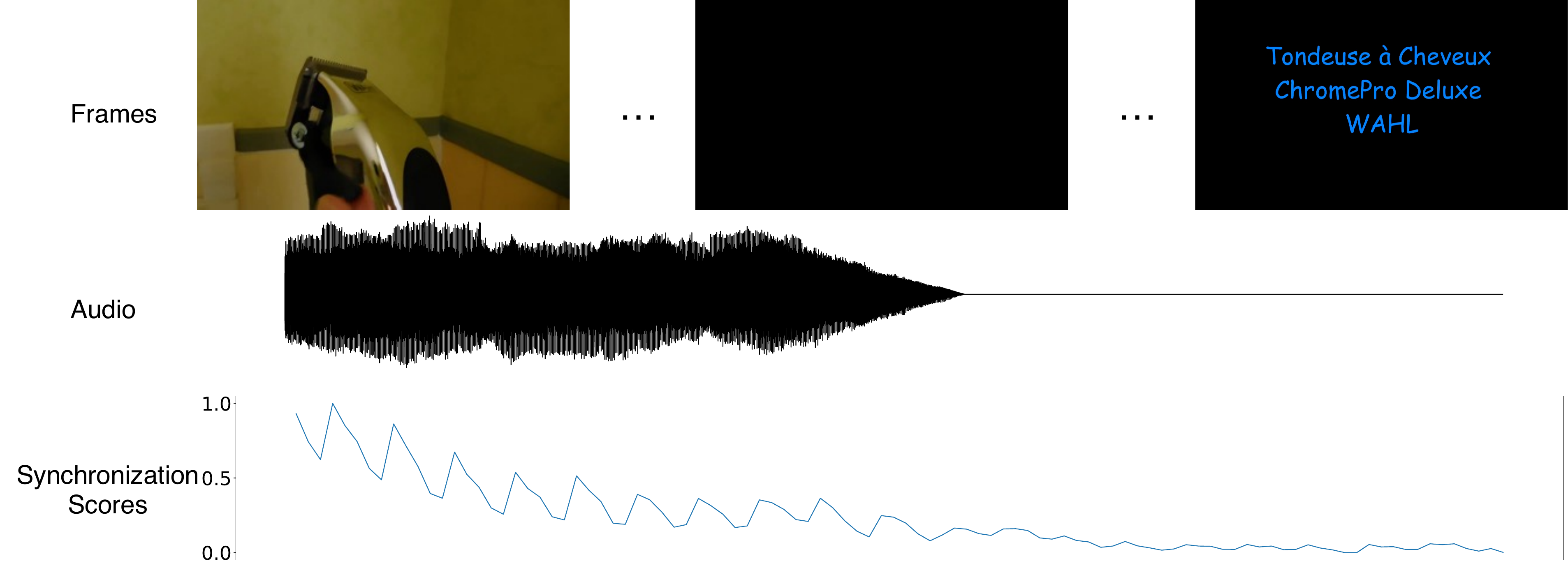}}\vspace{5mm}
\caption{Synchronisation scores along temporal axis. We analyse the hard cases which contain missing information of audio/video/both. }\label{fig:temp_visual_supp}
\end{figure*}

{\noindent \bf Visualising the temporal synchronisation.}
In order to understand the challenges in synchronising general sound,
we show video frames, audio and the model predicted synchronisation score from top to bottom, in Figure~\ref{fig:temp_visual_supp}.
Specifically, 
we input 6s of audio and the corresponding frames~(5 fps) with the same timestamp,
to generate a synchronisation score using our \dec~model. 
The bottom synchronisation score figure is then generated using a moving window with stride of 1 frame. Each video is 10s long in Figure~\ref{fig:temp_visual_supp}.

Ideally, we would like to see the scores being always around 1.0, 
however, in practise, 
synchronisation of general sound can be challenging due to the following reasons:
1), missing audio information, as shown in Figure~\ref{fig:a}, 
the bull makes sound only for the first few seconds, 
resulting in weak synchronisation in the later period. 
2), missing visual information, in Figure~\ref{fig:b}, 
the violin player is not present for the entire clip, 
causing low synchronisation scores. 
3), missing information of both modalities. 
The last example in Figure~\ref{fig:c} contains random frames with no sound;
this severely decreases the synchronisation scores. 
These hard cases suggest that experimenting on even longer input sequences may be beneficial, we therefore plan to investigate this in future work.

\clearpage
\section{Attention heatmaps visualisation}\label{sec:appendixD}
\begin{figure*}
\includegraphics[width=1\textwidth]{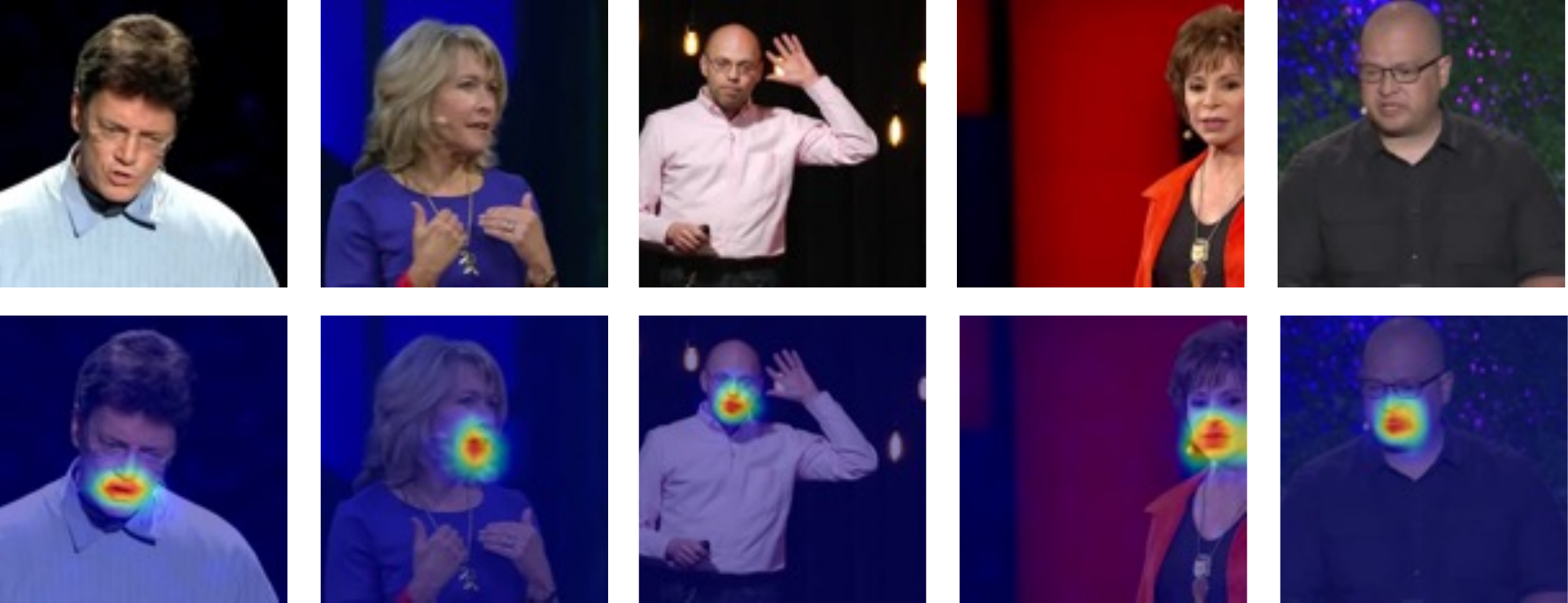}\hfill
\vspace{10mm}
\includegraphics[width=1\textwidth]{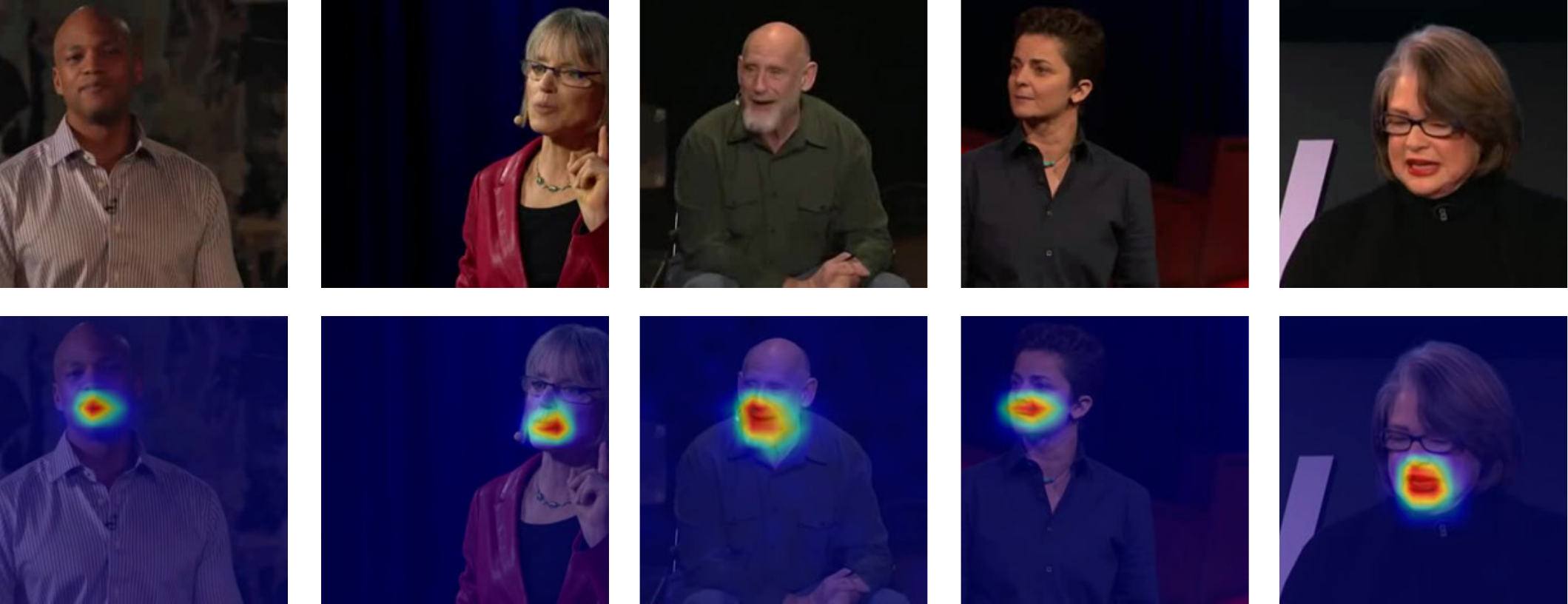}\hfill
\vspace{10mm}
\includegraphics[width=1\textwidth]{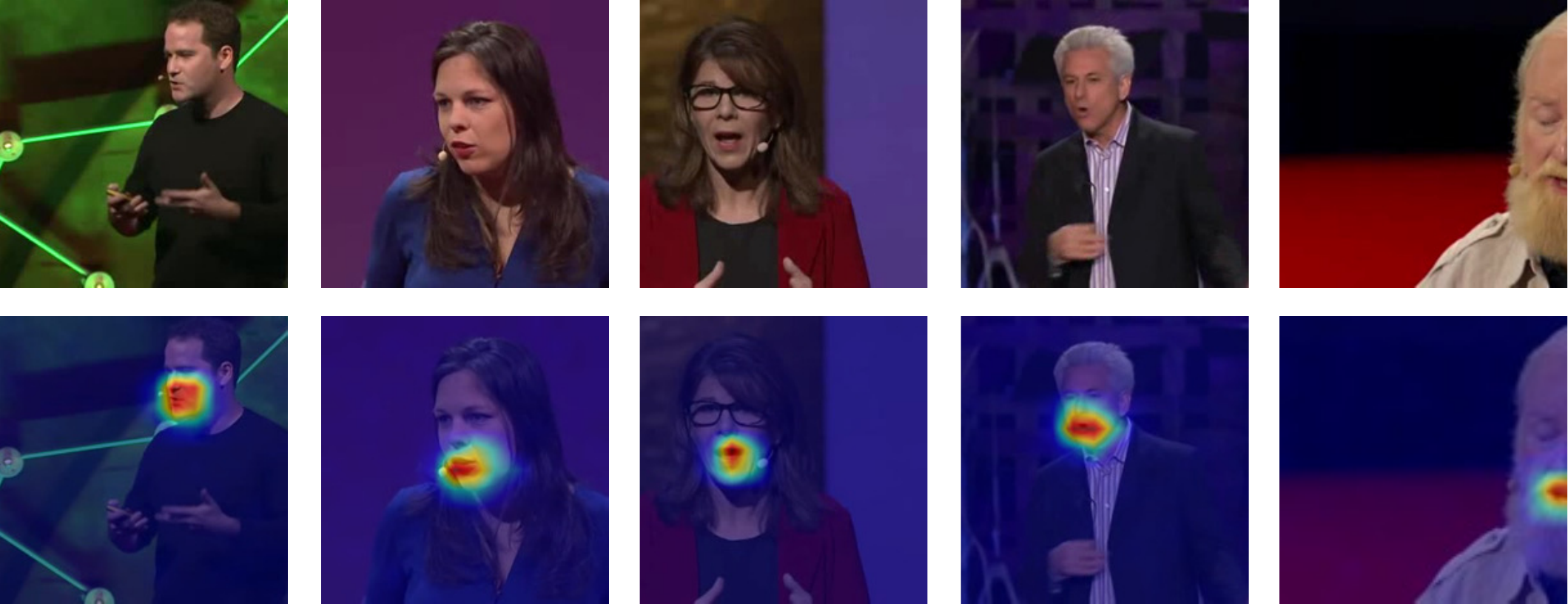}\hfill
\vspace{10mm}
\caption{Attention heatmap visualisations in LRS3 dataset}\label{fig:lrs_visual_supp}
\end{figure*}

\begin{figure*}
\includegraphics[width=1\textwidth]{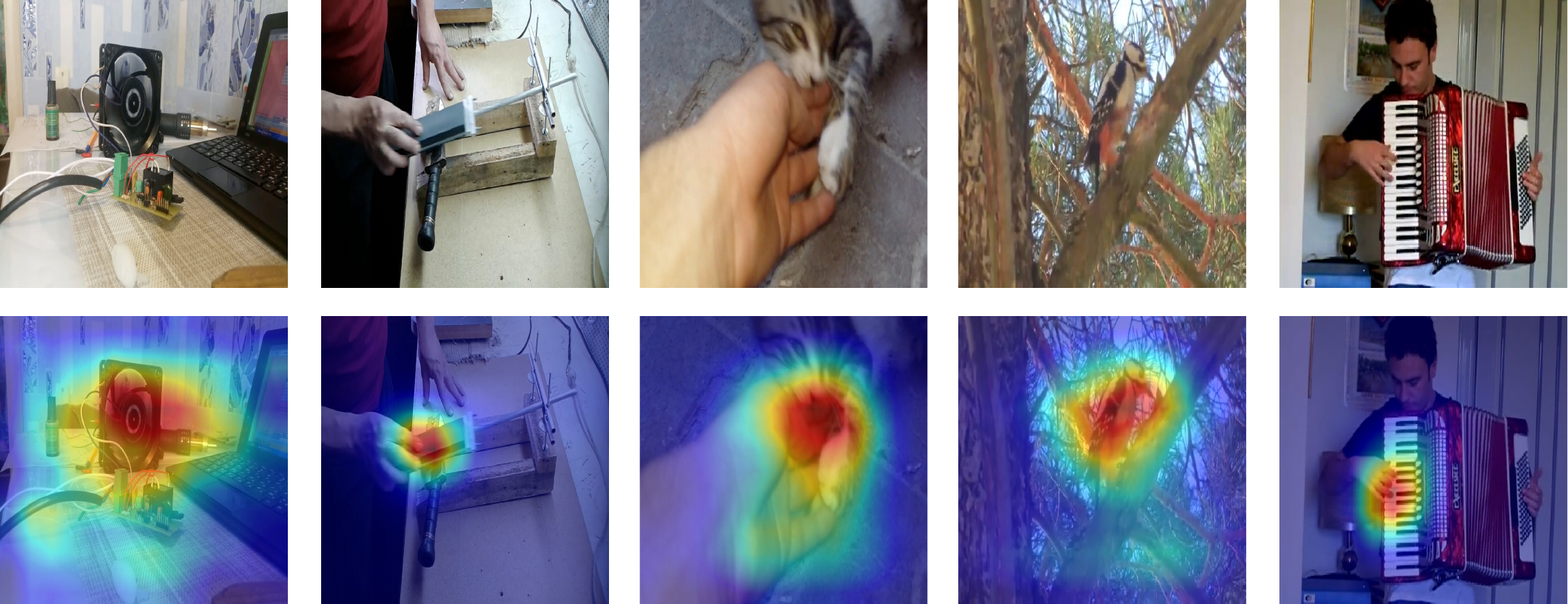}\hfill
\vspace{3mm}
\includegraphics[width=1\textwidth]{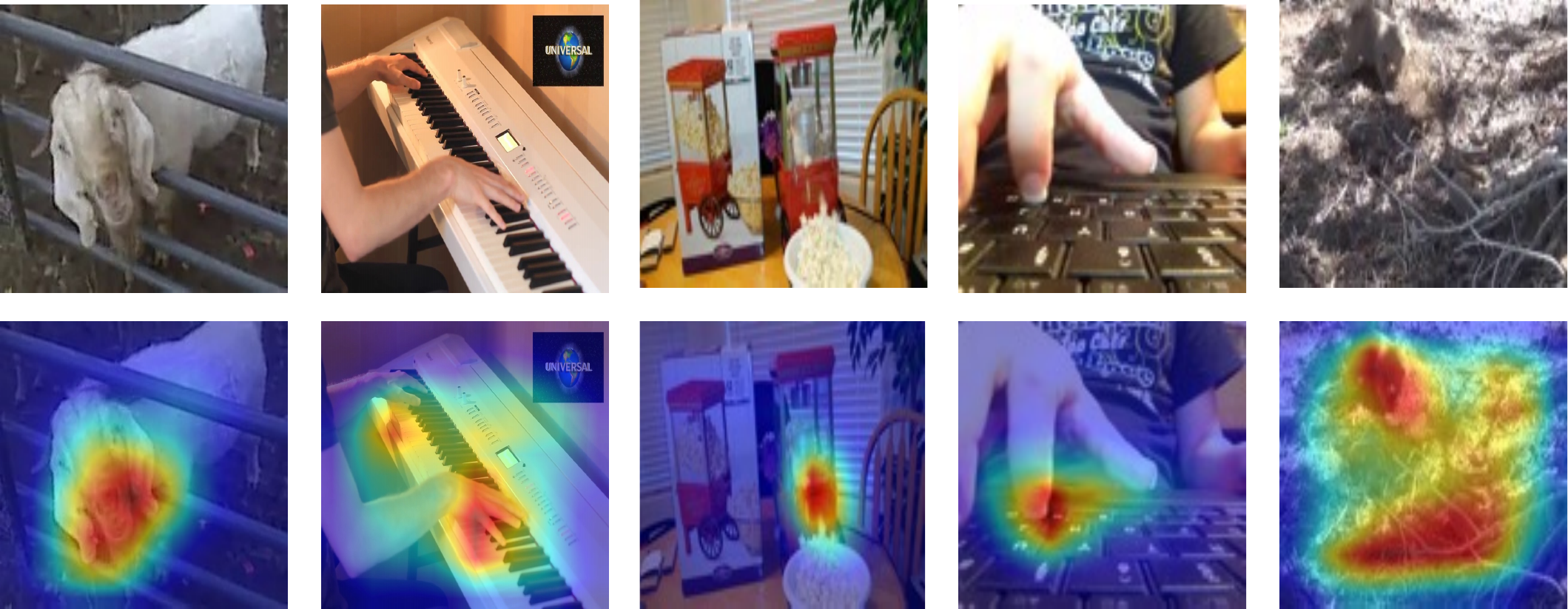}\hfill
\vspace{3mm}
\includegraphics[width=1\textwidth]{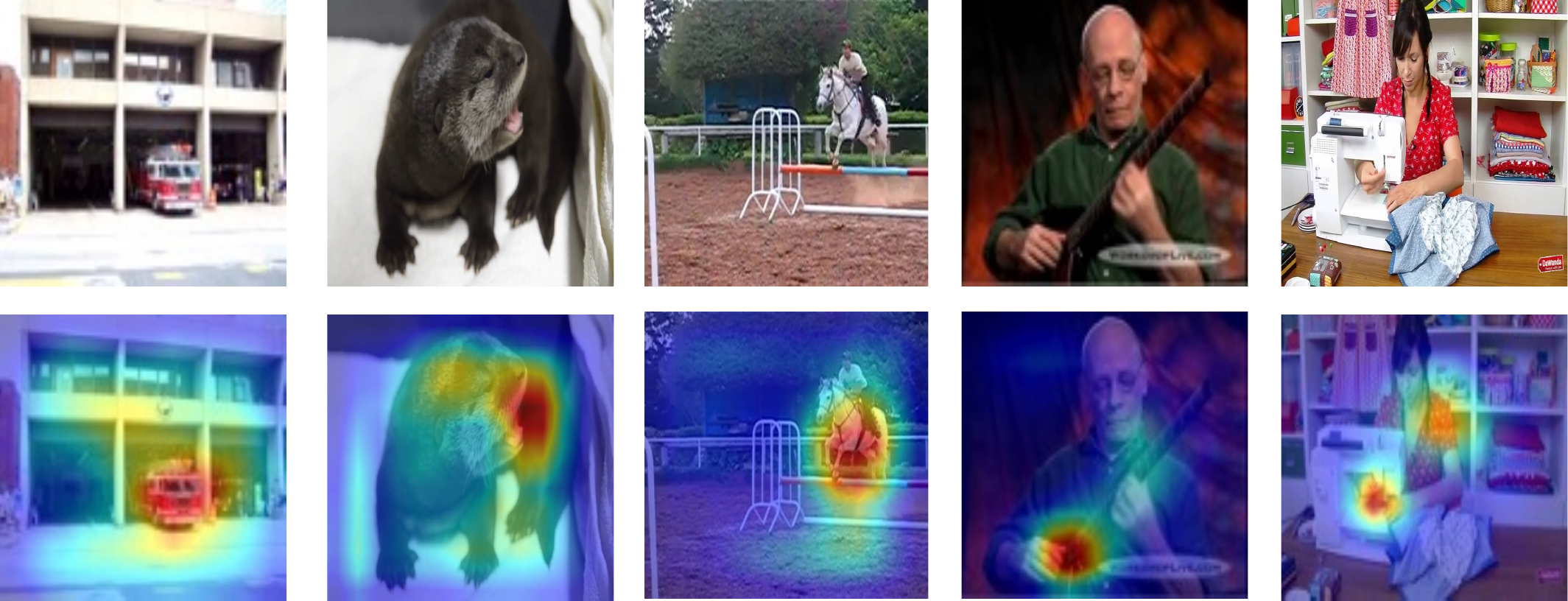}\hfill
\vspace{5mm}
\caption{Attention heatmap visualisations in VGG-Sound Sync dataset}\label{fig:vgg_visual_supp}
\end{figure*}

We show more image heatmap results from LRS3 in Figure~\ref{fig:lrs_visual_supp} and results from VGG-Sound Sync in Figure~\ref{fig:vgg_visual_supp}. Both models are trained using \dec~model.
The heatmaps are created by normalizing the attention matrix between the audio feature and visual feature in \dec~model.
In Figure~\ref{fig:lrs_visual_supp}, we can accurately localise the human mouth. In Figure~\ref{fig:vgg_visual_supp}, we localise the sound source in various classes, \eg~running electric fan, sharpen knife, cat meowing, etc. Note, rather than localise the entire sounding object, \eg piano, sewing machine, our model focus on the interaction points between the objects, \eg hands, sewing machine needle.



\clearpage
\noindent\textbf{Class list}
\begin{multicols}{2}
\begin{enumerate}
\item female singing
\item child singing
\item cutting hair with electric trimmers
\item machine gun shooting
\item female speech, woman speaking
\item playing steel guitar, slide guitar
\item male singing
\item people hiccup
\item people belly laughing
\item running electric fan
\item male speech, man speaking
\item people whispering
\item playing acoustic guitar
\item fox barking
\item people sniggering
\item people farting
\item dog whimpering
\item dog barking
\item people burping
\item people eating crisps
\item people sneezing
\item toilet flushing
\item people gargling
\item child speech, kid speaking
\item people humming
\item people babbling
\item train whistling
\item playing trumpet
\item mouse squeaking
\item dog bow-wow
\item people booing
\item alarm clock ringing
\item playing piano
\item children shouting
\item baby crying
\item dog growling
\item typing on typewriter
\item woodpecker pecking tree
\item playing banjo
\item shot football
\item people sobbing
\item cat growling
\item typing on computer keyboard
\item opening or closing car electric windows
\item forging swords
\item parrot talking
\item cat caterwauling
\item turkey gobbling
\item missile launch
\item cat meowing
\item firing cannon
\item bowling impact
\item lions roaring
\item lip smacking
\item striking bowling
\item people finger snapping
\item horse neighing
\item bull bellowing
\item people screaming
\item bathroom ventilation fan running
\item people slurping
\item cattle, bovinae cowbell
\item people slapping
\item skateboarding
\item hair dryer drying
\item playing accordion
\item skidding
\item ripping paper
\item race car, auto racing
\item playing volleyball
\item bird squawking
\item playing electronic organ
\item alligators, crocodiles hissing
\item playing bass drum
\item electric grinder grinding
\item duck quacking
\item goat bleating
\item chicken crowing
\item cap gun shooting
\item door slamming
\item playing harmonica
\item plastic bottle crushing
\item people running
\item sea lion barking
\item eletric blender running
\item people whistling
\item car passing by
\item playing bass guitar
\item firing muskets
\item popping popcorn
\item people coughing
\item dinosaurs bellowing
\item playing table tennis
\item playing tennis
\item rowboat, canoe, kayak rowing
\item playing snare drum
\item chopping food
\item cell phone buzzing
\item vehicle horn, car horn, honking
\item bee, wasp, etc. buzzing
\item playing cello
\item sharpen knife
\item playing french horn
\item sliding door
\item hammering nails
\item pigeon, dove cooing
\item opening or closing car doors
\item baltimore oriole calling
\item otter growling
\item penguins braying
\item people nose blowing
\item playing drum kit
\item frog croaking
\item people cheering
\item horse clip-clop
\item chipmunk chirping
\item elephant trumpeting
\item eagle screaming
\item people eating apple
\item cheetah chirrup
\item people giggling
\item playing trombone
\item chainsawing trees
\item snake hissing
\item cat purring
\item cat hissing
\item golf driving
\item playing hammond organ
\item playing violin, fiddle
\item using sewing machines
\item playing harpsichord
\item playing oboe
\item baby babbling
\item people shuffling
\item mouse clicking
\item swimming
\item people clapping
\item playing electric guitar
\item driving motorcycle
\item mouse pattering
\item crow cawing
\item tapping guitar
\item helicopter
\item rapping
\item bird chirping, tweeting
\item playing flute
\item tractor digging
\item skiing
\item francolin calling
\item lions growling
\item goose honking
\item bird wings flapping
\item car engine idling
\item train horning
\item playing harp
\item whale calling
\item chicken clucking
\item owl hooting
\item wind chime
\item snake rattling
\end{enumerate}
\label{itm:1}
\end{multicols}

\end{appendices}

\end{document}